\documentclass[conference]{IEEEtran}
\IEEEoverridecommandlockouts
\usepackage{comment}
\usepackage{cite}
\usepackage{makecell}
\usepackage{amsmath,amssymb,amsfonts}
\usepackage{algorithmic}
\usepackage{graphicx}
\usepackage{booktabs}
\usepackage{textcomp}
\usepackage{tabularx}

\usepackage{mathptmx} 
\usepackage[whole]{bxcjkjatype} 
\usepackage[unicode,hidelinks]{hyperref} 

\newcolumntype{C}{>{\centering\arraybackslash}X} 
\usepackage{xcolor}
\def\BibTeX{{\rm B\kern-.05em{\sc i\kern-.025em b}\kern-.08em
    T\kern-.1667em\lower.7ex\hbox{E}\kern-.125emX}}

\usepackage[normalem]{ulem}
\newif\ifrevision
\revisionfalse
\ifrevision
  \newcommand{\rev}[1]{\textcolor{red}{#1}}
  \newcommand{\del}[1]{\textcolor{gray}{\sout{#1}}}
\else
  \newcommand{\rev}[1]{#1}
  \newcommand{\del}[1]{}
\fi

\begin{document}

\title{Dialect-Robust Speech Language Models \\ with Synthetic Pseudo-Dialect Augmentation}

\author{
\IEEEauthorblockN{Shunsuke Mitsumori\IEEEauthorrefmark{1}\IEEEauthorrefmark{2},
Tomoya Mizumoto\IEEEauthorrefmark{1},
Yusuke Fujita\IEEEauthorrefmark{1}}
\IEEEauthorblockA{\IEEEauthorrefmark{1}\textit{SB Intuitions},
\IEEEauthorrefmark{2}\textit{Waseda University}, Tokyo, Japan \\
shunsuke.mitsumori@sbintuitions.co.jp}
}


\maketitle

\begin{abstract}
Speech Language Model (SLM) performance often degrades on dialects due to data scarcity. Conventional text-to-speech (TTS) augmentation struggles to cover diverse dialects as it requires a certain amount of real dialect speech. We propose synthesizing pseudo-dialect speech by converting LLM-generated dialect text via a standard-language TTS model, requiring zero real dialect speech. Additionally, we introduce intermediate standard-text prediction during training, acting as semantic normalization for downstream tasks.
\del{To evaluate dialect understanding capabilities via speech translation from dialects to English, we evaluate our framework across Japanese, German, and Chinese dialects. The proposed framework consistently improves BLEU scores over the baselines, increasing from 24.21 to 28.26 for Japanese, 22.37 to 32.47 for German, and 11.61 to 16.37 for Chinese.}
\rev{We evaluate dialect understanding via dialect-to-English speech translation across Japanese, German, and Chinese dialects. Compared to synthetic standard speech baselines, pseudo-dialect augmentation improves scores for Japanese (from 25.38 to 26.24) and German (from 31.57 to 32.47). Furthermore, the intermediate standard-text prediction effectively bridges the semantic gap, boosting performance to 28.26 for Japanese and from 11.67 to 16.37 for Chinese.}
These results suggest that our approach scales to various languages without requiring speech resources specific to each dialect.
\end{abstract}


\section{Introduction}

Dialect speech possesses unique linguistic features (e.g., distinct vocabulary and grammar) and acoustic features that differ significantly from the standard language~\cite{Chambers_Trudgill_1998,Zampieri_Nakov_2021}\rev{.} 
Because of these discrepancies, conventional speech architectures have historically struggled with dialect robustness~\cite{miwa23_interspeech, li2018_icassp, yoo2019highly}.
Even recent advancements in Speech Language Model (SLM)~\cite{cui-etal-2025-recent,Peng2024ASO}, which leverage LLMs to achieve remarkable success in tasks such as speech recognition, translation, and dialogue~\cite{zhang-etal-2023-speechgpt,defossez2024moshispeechtextfoundationmodel,fang2025llamaomni,chen-etal-2025-slam}, face severe performance degradation when encountering dialects~\cite{mizumoto2025evaluating,chen-etal-2026-voicebench}.
The degradation is primarily attributed to the lack of real dialect speech data for model training.

In the field of low-resource languages, data augmentation approaches using speech synthesis have been adopted to address such data scarcity. The primary approach is to convert text obtained via LLM or web collection into speech with a Text-to-Speech (TTS) model trained on real speech of the target language~\cite{wang-etal-2025-tens,dao25_interspeech}. 
However, this augmentation strategy cannot cover various dialects because it requires \rev{a} sufficient amount of real dialect speech for training TTS models. 

\begin{figure*}[t]
    \centering
    \includegraphics[width=0.9\linewidth]{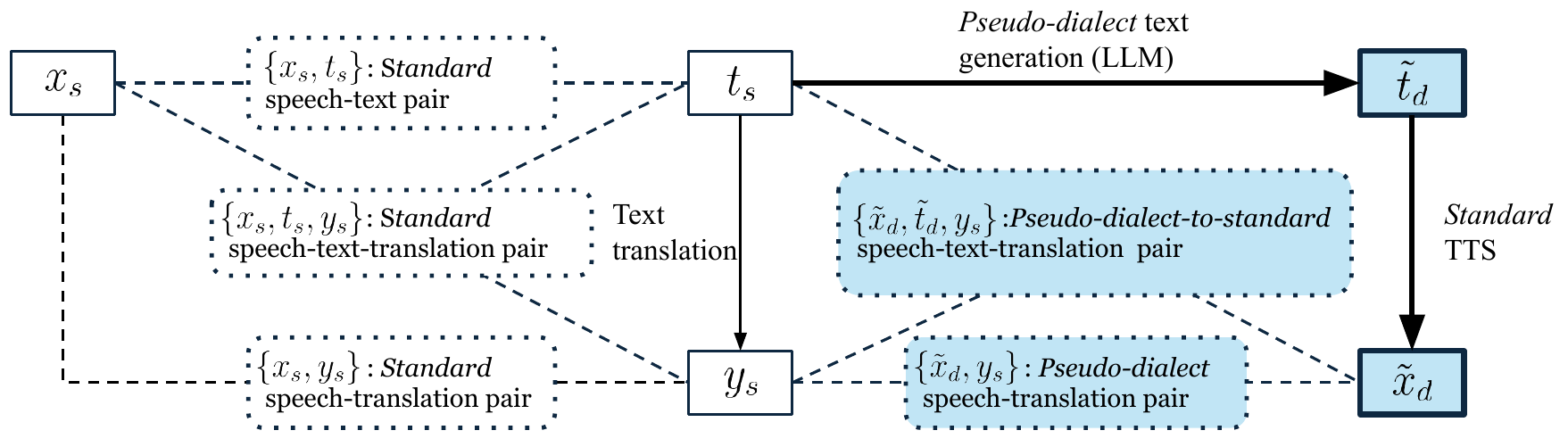}
    \vspace{0mm}
    \caption{Proposed pseudo-dialect speech augmentation. From a standard pair $\{x_s, t_s\}$ and translation $y_s$, an LLM and standard TTS model generate pseudo-dialect text $\tilde{t}_d$ and speech $\tilde{x}_d$. This yields $\{\tilde{x}_d, y_s\}$ and $\{\tilde{x}_d, t_s, y_s\}$ for single-task and multi-task SLM training, respectively.}
    \label{fig:method}
    \vspace{-2mm}
\end{figure*}

To expand dialect coverage without relying on real dialect audio, we focus on the potential of reusing standard-language speech resources. 
Since dialects are varieties of a standard language rather than completely distinct languages~\cite{Chambers_Trudgill_1998}, they inherently share overlapping acoustic features. 
Leveraging this acoustic similarity, we propose a novel dialect adaptation framework that bypasses the need for dialect-specific TTS models by utilizing ``pseudo-dialect'' speech. 
Specifically, we first employ an LLM to rewrite standard text into text with dialectal linguistic features. 
Then, we utilize a standard-language TTS model to convert this augmented text into pseudo-dialect speech. 
Since this framework requires zero real dialect speech resources, it enables data augmentation for arbitrary dialects.


Furthermore, leveraging the property that pseudo-dialect speech inherently pairs with standard language text during its generation, we introduce multi-task learning with intermediate standard-text prediction. We exploit this standard text as an explicit intermediate output before predicting the final target task. By projecting the input into the standard language, we force the model to map the dialectal audio into a rich semantic space where the backend LLM can perform strong reasoning.

We verified the general effectiveness of the proposed framework through two main perspectives. To measure dialect understanding capabilities of SLM, we followed previous \rev{work}~\cite{mizumoto2025evaluating} and evaluated the models using a dialect-to-English translation task. First, to test its effectiveness across languages, we evaluated SLMs across Japanese, German, and Chinese dialects. Second, to test architectural generalizability of our pseudo-dialect speech augmentation, we applied our augmentation to conventional end-to-end models (Whisper) and cascade models (Whisper+LLM) using Japanese data. 

Experimental results showed that\rev{, compared to a synthetic standard speech baseline,} the pseudo-dialect augmentation improved English translation BLEU scores for Japanese (from \rev{25.38} to 26.24) and German (from \rev{31.57} to 32.47). Furthermore, intermediate standard-text prediction boosted performance for Japanese (to 28.26) and Chinese (from 11.67 to 16.37). Regarding architectural generalizability, the pseudo-dialect speech augmentation also benefited the direct Whisper model (from \rev{18.44} to 18.98) and the cascade system (from \rev{29.65} to 30.44) \rev{over their respective synthetic standard speech baselines.}
\del{These results confirm that our proposed pseudo-dialect speech augmentation is broadly effective across various languages and model architectures, and that intermediate standard-text prediction serves as a robust mechanism for SLMs.}

\section{Proposed Method} \label{method}
In this section, we formulate the proposed dialect adaptation framework. Our target task is to predict a translation text $y$ from an input speech signal $x$ by modeling $p(y|x)$. In practice, directly training $p(y|x)$ is challenging because paired data $\{(x, y)\}$ are scarce. To obtain pseudo translation labels, we start from an automatic speech recognition (ASR) corpus of standard speech-text pairs $\{(x_s, t_s)\}$ and apply a text translation model $p_\theta(y|t_s)$ to the transcripts:
\begin{equation}
y_s \sim p_\theta(y|t_s).
\end{equation}
This yields pseudo-labeled speech translation data $\{(x_s, y_s)\}$. However, the available ASR corpus mostly consists of standard speech $x_s$, and the lack of dialect speech $x_d$ is a fundamental bottleneck. 
To address this, we propose a dialect adaptation framework using pseudo-dialect speech augmentation (Figure~\ref{fig:method}) and multi-task learning with intermediate standard-text prediction. 

\subsection{Pseudo-Dialect Speech Augmentation}
To mitigate the shortage of dialect speech, we synthesize pseudo-dialect training examples without relying on real dialect speech. Given a standard text $t_s$, we first generate a pseudo-dialect text $\tilde{t}_d$ using an LLM-based dialect generation model $p_\phi(\tilde{t}_d|t_s)$:
\begin{equation}
    \tilde{t}_d \sim p_\phi(\tilde{t}_d|t_s).
\end{equation}
Next, we synthesize pseudo-dialect speech $\tilde{x}_d$ from $\tilde{t}_d$ using a standard TTS model $p_\psi(x|t)$:
\begin{equation}
    \tilde{x}_d \sim p_\psi(x|\tilde{t}_d).
\end{equation}
Finally, we pair the synthesized speech $\tilde{x}_d$ with the pseudo translation label $y_s$ to obtain $\{(\tilde{x}_d, y_s)\}$ for SLM training.
Note that the TTS model $p_\psi(x|t)$ is trained on standard speech-text pairs, and thus $\tilde{x}_d$ is typically perceived as \textit{non-native-sounding} dialect speech. Nevertheless, because $\tilde{t}_d$ contains dialectal lexical and syntactic cues, $\tilde{x}_d$ provides useful supervision for dialect robustness.

\subsection{Intermediate Standard-Text Prediction for SLM}

With the proposed augmentation, each training sample forms a triplet $(\tilde{x}_d, t_s, y_s)$. To effectively translate dialect speech $x_d$, we exploit the standard text $t_s$ as an explicit intermediate representation. By projecting the input into $t_s$, we map the audio into a rich semantic space where the underlying LLM possesses the strongest semantic reasoning capabilities. This process acts as a semantic denoising step, effectively bridging the dialectal input and the target translation. We factorize the speech translation model as:
\begin{equation}
    \label{eq:marginalization}
    p(y|x) = \sum_{t_s} p(y|t_s\rev{,x})p(t_s|x).
\end{equation}
In practice, we approximate the marginalization over $t_s$ by decoding the most probable sequence. This yields a chain-of-thought process that predicts the standard text $t_s$ as a semantic bridge followed by the final translation $y$. During training, the model is jointly optimized using both the pseudo-dialect data $(\tilde{x}_d, t_s, y_s)$ and the original standard data $(x_s, t_s, y_s)$.


\begin{table*}[t]
    \centering
    \caption{Output formats for each multi-task learning configuration. Variables follow the notation introduced in Section 2, where the transcription target ($\tilde{t}_d$ or $t_s$) depends on the input speech.}
    \vspace{0mm}
    \label{tab:multitask_formats}
    \small
    
    \setlength{\tabcolsep}{4pt}
    
    \begin{tabular}{l ccc l} \hline
        \rev{Output Targets} & \makecell{Transcription\\($\tilde{t}_d$ or $t_s$)} & \makecell{Standard text\\($t_s$)} & \makecell{Translation text\\($y$)} & Output Format \\ \hline
        \rev{ASR+English} & \checkmark &            & \checkmark & \texttt{<ASR>}$\tilde{t}_d$ or $t_s$\texttt{</ASR><en>}$y$\texttt{</en>} \\
        \rev{Standard+English} &            & \checkmark & \checkmark & \texttt{<standard>}$t_s$\texttt{</standard><en>}$y$\texttt{</en>} \\
        \rev{ASR+Standard+English} & \checkmark & \checkmark & \checkmark & \texttt{<ASR>}$\tilde{t}_d$ or $t_s$\texttt{</ASR><standard>}$t_s$\texttt{</standard><en>}$y$\texttt{</en>} \\ \hline
    \end{tabular}
    \vspace{-2mm}
\end{table*}

\section{Experimental Setup}\label{experiment}
We verify the effectiveness of the pseudo-dialect speech augmentation and intermediate standard-text prediction, as proposed in Section \ref{method}, within the SLM framework. Because the linguistic relationship between regional dialects and the standard language varies across different languages, we conduct our experiments on Japanese, Chinese, and German to evaluate the general effectiveness of our approach.
\subsection{Evaluation Tasks}
To evaluate dialect understanding, we focus on the English translation task taking real dialect speech as input. This task verifies whether the model truly understands the underlying semantics of the dialect speech to predict the final translation ($y$)~\cite{mizumoto2025evaluating}. 
To evaluate the English translation performance, BLEU~\cite{Papineni-2002-bleu} scores were computed using SacreBLEU~\cite{post-2018-call}.

\subsection{Model and Training Setup}\label{setup}
In this study, we constructed SLMs connecting a speech encoder and an LLM with a projector. Across all languages, we adopted Whisper large-v3~\cite{radford2022whisper} as the speech encoder. For the LLM backend, we utilized Llama-3.1-Swallow-8B-Instruct-v0.3~\cite{Fujii:COLM2024} for Japanese, and Llama-3.1-8B-Instruct~\cite{llama3modelcard} for German and Chinese. The projector consists of two 1-dimensional convolution layers (kernel size 4, stride 2) with GELU activations and a final linear layer. Following previous studies on SLMs~\cite{xiong2024freeze,du2024cot,ma2025speech,yang2024ctc}, the parameters of the speech encoder and the LLM were frozen during training, while only the projector parameters were updated.

All SLM models were trained on 8 NVIDIA H100 (80GB) GPUs for 10 epochs using the AdamW optimizer. The effective batch size was set to 512, and the learning rate peaked at 1e-4. The SLM training times varied depending on the dataset sizes of each language. For Japanese, training took approximately 2 days without pseudo-dialect speech augmentation, and 4 days with the augmentation. For Chinese, it required slightly over 1 day without augmentation and approximately 2.5 days with it. For German, training took approximately 6 hours and 12 hours, respectively.

\begin{table*}[t]
  \caption{Comparison of dialect understanding capabilities evaluated via dialect-to-English speech translation performance (BLEU) across multiple languages, data augmentation settings, and task output formats. For the output targets, ``English'' denotes direct translation, while ``ASR'' and ``Standard'' represent intermediate speech transcription and standard-text prediction, respectively.}
  \vspace{-0mm}
  \label{tab:multilingual}
  \small
  \centering
  \begin{tabular}{llccc}
    \toprule
    \textbf{Synthetic data added} & \textbf{Output Target} & \textbf{Japanese} & \textbf{German} & \textbf{Chinese} \\
    \midrule
    No Augmentation (baseline) & English & 24.21 & 22.37 & 11.61 \\
    Synthetic standard speech & English & 25.38 & 31.57 & 11.72 \\
    Pseudo-dialect speech & English & 26.24 & \textbf{32.47} & 11.67 \\
    Pseudo-dialect speech & ASR+English & 27.67 & 29.81 & 16.30 \\
    Pseudo-dialect speech & Standard+English & \textbf{28.26} & 30.73 & \textbf{16.37} \\
    Pseudo-dialect speech & ASR+Standard+English & 24.95 & 29.92 & 13.24 \\
    \bottomrule
  \end{tabular}
  \vspace{-2mm}
\end{table*}

\subsection{Datasets}\label{dataset}
\subsubsection{Evaluation Datasets}\label{evaluate}
To evaluate dialect understanding, we utilized corpora containing real speech from regional dialects. For Japanese, we used CPJD~\cite{takamichi-saruwatari-2018-cpjd} comprising real speech from 20 dialects. For German, we used SwissDial~\cite{doganschonberger2021swissdialparallelmultidialectalcorpus} consisting of 8 dialects. For Chinese, we utilized KeSpeech~\cite{Tang2021KeSpeechAO} containing 8 dialect subsets. To ensure accurate evaluation across all three languages, we generated high-quality English reference labels from their respective standard language parallel texts using GPT-4o. These real dialect speeches were strictly excluded from the training data.

\subsubsection{Training Datasets}
Depending on the experimental configurations, models for each language were trained using real standard language resources, either alone, augmented with synthesized pseudo-dialect speech, or augmented with synthetic standard speech (which is included for comparison purposes as detailed in Section \ref{config}).

For Japanese, we utilized ReazonSpeech large v2 comprising approximately 2.6M utterances~\cite{reazonspeech2023}, employing its transcripts as standard Japanese labels and as the source text for the data augmentation pipeline. Additionally, we included real standard speech from Speech-BSD (20,000 utterances)~\cite{shimizu-etal-2023-towards} and the Japanese subset of CoVoST2 (1,119 utterances)~\cite{wang2020covost}, along with their human-annotated Japanese transcripts and English translations.

For German, we used the German subset of Multilingual LibriSpeech~\cite{pratap20_interspeech} comprising approximately 0.4M utterances. For Chinese, we utilized a subset of 3M utterances from WenetSpeech~\cite{zhang2022wenetspeech}. These datasets served as standard speech training data and the source texts for the data augmentation pipeline.

Because ReazonSpeech, the German subset of Multilingual LibriSpeech, and WenetSpeech lack English translations, we \del{utilized Qwen2.5-32B-Instruct to efficiently generate English labels for these large-scale datasets using their standard text transcripts.} \rev{generated their English labels using Qwen2.5-32B-Instruct~\cite{qwen2025qwen25technicalreport}. We chose a different model from the LLM used to generate the English references for evaluation (GPT-4o) to prevent overfitting to the specific stylistic biases of a single machine translation system, following previous work~\cite{mizumoto2025evaluating}.}

\begin{table*}[t]
    \centering
    \caption{Comparison of English translation performance (BLEU) across different types of data augmentation, showing the average score and individual results for 20 Japanese dialects.}
    \label{tab:data_augmentation_results}
    \small
    \vspace{-2mm}
    \setlength{\tabcolsep}{2pt}
    
    \begin{tabularx}{\textwidth}{l|c|CCCCCCCCCC} \hline
        Synthetic data added & Avg. & Akita & Hiroshima & Iyo & Fukuoka & Awa & Hokkaido & Fukui & Kyoto & Miyazaki & Enshu \\ \hline
        No Augmentation (baseline) & 24.21 & 19.10 & 25.97 & 27.49 & 26.72 & 25.34 & 30.19 & 27.69 & 24.35 & 21.73 & 23.83 \\
        Synthetic standard speech & 25.38 & 19.90 & 26.67 & 27.58 & 28.46 & 26.29 & 30.17 & 29.57 & 25.97 & 23.23 & 24.70 \\
        Pseudo-dialect speech & \textbf{26.24} & \textbf{20.12} & \textbf{27.39} & \textbf{29.16} & \textbf{30.39} & \textbf{27.23} & \textbf{30.78} & \textbf{30.13} & \textbf{27.14} & \textbf{24.98} & \textbf{25.22} \\ \hline
    \end{tabularx}
    
    \vspace{3mm}
    
    \begin{tabularx}{\textwidth}{l|c|CCCCCCCCCC} \hline
        Synthetic data added & Avg. & Okayama & Saitama & Tosa & Morokata & Tsugaru & Iwaki & Izumo & Kanazawa & Nara & Osaka \\ \hline
        No Augmentation (baseline) & 24.21 & 24.83 & 28.66 & 25.36 & 10.60 & 14.30 & 27.75 & 22.55 & \textbf{21.68} & 26.67 & 28.19 \\
        Synthetic standard speech & 25.38 & 25.96 & 31.03 & 26.78 & 10.12 & 14.65 & \textbf{30.12} & \textbf{24.92} & 20.04 & \textbf{28.81} & \textbf{29.58} \\
        Pseudo-dialect speech & \textbf{26.24} & \textbf{28.97} & \textbf{31.98} & \textbf{28.14} & \textbf{12.41} & \textbf{16.70} & 30.05 & 23.28 & 21.45 & 28.40 & 29.45 \\ \hline
    \end{tabularx}
    \vspace{-2mm}
\end{table*}

\begin{table*}[t]
    \centering
    \caption{Impact of intermediate prediction on English translation (BLEU), showing the average score and individual results for 20
Japanese dialects. All models were trained on the dataset augmented with pseudo-dialect speech.}
    \label{tab:multitask_results}
    \small
    \vspace{-2mm}
    \setlength{\tabcolsep}{2pt}
    
    \begin{tabularx}{\textwidth}{l|c|CCCCCCCCCC} \hline
        Output Targets & Avg. & Akita & Hiroshima & Iyo & Fukuoka & Awa & Hokkaido & Fukui & Kyoto & Miyazaki & Enshu \\ \hline
        English (Single-task) & 26.24 & 20.12 & 27.39 & 29.16 & 30.39 & 27.23 & 30.78 & 30.13 & 27.14 & 24.98 & 25.22 \\
        \rev{ASR}+English & 27.67 & \textbf{23.26} & 30.24 & 31.27 & \textbf{33.50} & 28.58 & 30.96 & \textbf{31.51} & 27.94 & \textbf{26.96} & 26.29 \\
        Standard+English & \textbf{28.26} & 22.20 & \textbf{32.53} & \textbf{31.68} & 32.25 & \textbf{29.72} & \textbf{31.21} & 30.27 & \textbf{29.88} & 26.93 & \textbf{26.39} \\
        ASR+Standard+English & 24.95 & 19.23 & 25.12 & 25.93 & 29.18 & 27.13 & 27.45 & 29.42 & 26.46 & 22.22 & 25.36 \\ \hline
    \end{tabularx}
    
    \vspace{2mm}
    
    \begin{tabularx}{\textwidth}{l|c|CCCCCCCCCC} \hline
        Output Targets & Avg. & Okayama & Saitama & Tosa & Morokata & Tsugaru & Iwaki & Izumo & Kanazawa & Nara & Osaka \\ \hline
        English (Single-task) & 26.24 & 28.97 & 31.98 & 28.14 & 12.41 & 16.70 & 30.05 & 23.28 & 21.45 & 28.40 & 29.45 \\
        \rev{ASR}+English & 27.67 & 29.59 & \textbf{34.09} & 28.32 & 11.48 & 17.65 & 30.12 & 24.83 & 23.63 & 30.38 & \textbf{30.95} \\
        Standard+English & \textbf{28.26} & \textbf{29.79} & 34.00 & \textbf{29.22} & 12.34 & \textbf{20.43} & \textbf{31.20} & \textbf{27.25} & \textbf{24.63} & \textbf{30.62} & 30.85 \\
        ASR+Standard+English & 24.95 & 25.05 & 29.04 & 25.45 & \textbf{12.54} & 16.68 & 30.12 & 22.53 & 21.96 & 27.45 & 28.45 \\ \hline
    \end{tabularx}
    \vspace{-2mm}
\end{table*}

\subsection{Experimental Configurations}

\subsubsection{Data Augmentation Configurations}\label{config}

We compare three training data settings to evaluate the proposed pseudo-dialect speech. All models in this comparison are trained as standard single-task models.

\vspace{1mm}
\textbf{No Augmentation:} Uses only the real speech data from Section~\ref{dataset}, serving as a baseline without synthetic data.
This original training set consists of approximately 2.6M utterances for Japanese, 0.4M for German, and 3.0M for Chinese.

\vspace{1mm}
\textbf{Synthetic standard speech:} 
Adds synthesized standard utterances generated from the respective standard source texts described in Section~\ref{dataset}.
By including a 1:1 match of synthesized utterances, this configuration doubles the training volume to a total of 5.2M for Japanese, 0.8M for German, and 6.0M for Chinese.
For Japanese, we used Tsukasa-Speech~\cite{TsukasaSpeech2024} with reference speech from the JVS corpus~\cite{takamichi2019jvscorpusfreejapanese}.
For German and Chinese, Qwen-TTS~\cite{qwen2025qwen25technicalreport} was utilized to convert the standard transcripts using the original standard speech as reference.
This setup isolates the effect of pure data volume, verifying that performance gains come specifically from dialectal linguistic features.

\vspace{1mm}
\textbf{Pseudo-dialect speech:} Adds the proposed pseudo-dialect speech \rev{to the baseline dataset}.
The standard source transcripts for each language were translated into one of the target regional dialects chosen at random (20 dialects for Japanese, 8 for German, and 8 for Chinese).
This dialectal rewriting was performed via gpt-oss-120b~\cite{openai2025gptoss120bgptoss20bmodel} across all languages.
The rewritten texts were then synthesized into speech using the identical TTS setup as the synthetic standard speech for each language.
To ensure a fair comparison, this configuration maintains the exact same total data volume as the synthetic standard speech setting.

\subsubsection{Multi-task Learning Configurations}
To verify the effectiveness of multi-task learning with intermediate standard-text prediction for the SLM, we trained models on the full dataset augmented with pseudo-dialect speech in Section~\ref{config}. The designated target output formats for these multi-task configurations are defined in Table \ref{tab:multitask_formats}. By generating this concatenated sequence, the joint probability formulated in Eq. \ref{eq:marginalization} is implicitly optimized through the standard autoregressive next-token prediction (cross-entropy) loss.
During evaluation of the multi-task models, the final English translation was obtained by extracting the text enclosed within the \texttt{<en>} and \texttt{</en>} tags using regular expressions.

\section{Experimental Results}\label{result}
Table~\ref{tab:multilingual} presents the overall dialect understanding performance evaluated via English translation BLEU on the real dialect speech from Section~\ref{evaluate}, across multiple languages under the various data augmentation settings and multi-task learning configurations described in Section~\ref{config}. The results indicate that our framework successfully improved the dialect understanding performance across all target languages. However, whether pseudo-dialect speech augmentation or multi-task learning primarily drove these improvements highly depended on the linguistic properties of each language.
\rev{To rigorously evaluate these performance differences, we conducted paired bootstrap resampling tests~\cite{koehn2004statistical} with 1,000 iterations ($p < 0.05$) at the overall language or individual dialect levels. Note that Japanese dialect subsets have small sample sizes ($\sim$250 utterances). Thus, non-significant results may stem from low statistical power rather than an absence of effect.}

\begin{table*}[t]
\centering
\caption{Qualitative analysis of translation outputs from the English translation task. Asterisk (*) and dagger (\dag) indicate failure cases due to unseen vocabulary and acoustic gaps.}
\vspace{-2mm}
\label{tab:qualitative_analysis}
\small
\setlength{\tabcolsep}{3pt} 
\begin{tabularx}{\textwidth}{l X X X X}
\toprule
\textbf{Category} & \textbf{Input Audio (Dialect)} & \textbf{English Ground Truth} & \textbf{Baseline Output} & \textbf{Proposed Output} \\
\midrule
\textbf{Improved 1} & ほんで食材も & And we also \textbf{ran out} & And I also \textbf{bought} & And we also \textbf{ran out of} \\
(Kyoto) & \textbf{たりひんかった}んどす． & \textbf{of} ingredients. & some ingredients. & ingredients. \\
\midrule
\textbf{Improved 2} & きっと素敵な休日に & \textbf{I'm sure it will be} a & \textbf{I hope you have} a & \textbf{I'm sure it will be} a \\
(Iwaki) & \textbf{なっぺよ}． & wonderful holiday. & wonderful holiday. & wonderful holiday. \\
\midrule
\midrule
\textbf{Failure 1} & \textbf{がめる}もの\textbf{がめる}もの， & The things I \textbf{pick up} & There are a lot of & There are a lot of \\
(Kanazawa)* & 古いものが多いがやて． & tend to be old. & old things here. & old things here. \\
\midrule
\textbf{Failure 2} & \textbf{赤}のスマホでも & A \textbf{red} smartphone & I might as well use & Maybe it's okay if \\
(Tsugaru)\dag & いいがもな． & might be nice, too. & my smartphone. & it's an iPhone. \\
\bottomrule
\end{tabularx}
\vspace{-2mm}
\end{table*}

\begin{table*}[t]
\centering
\caption{Comparison of English translation outputs across multi-task configurations. All models were trained on the identical dataset augmented with pseudo-dialect speech, differing only in their target output formats.}
\vspace{-2mm}
\label{tab:multitask_comparison}
\small
\setlength{\tabcolsep}{3pt} 
\begin{tabularx}{\textwidth}{l X}
\toprule
\textbf{Output \rev{Target}} & \textbf{Output Content} \\
\midrule
Ground Truth & 
\textbf{ASR:} 必要ねぇのにスマートフォンばもってる理由は、緊急連絡用に、親が\textbf{わさけだ}はんで。\newline
\textbf{Standard:} 必要がないのにスマートフォンをもっている理由は、緊急連絡用にと親が私に与えたから。\newline
\textbf{English:} The reason I have a smartphone even though I don't need one is because my parents gave it to me for emergency contact purposes. \\
\midrule
English (Single-task) & 
\textbf{English:} The reason I have a smartphone even though I don't need one is because my parents \textbf{insisted, saying it's for emergencies}. \\
\midrule
\rev{ASR}+English & 
\textbf{ASR:} 必要ねぇのにスマートフォンば持ってる理由は、緊急連絡用に、親が\textbf{話さけだ}はんで。\newline 
\textbf{English:} The reason for carrying a smartphone even though it's not necessary is because it's for emergency contact, and \textbf{parents can't talk}. \\
\midrule
Standard+English & 
\textbf{Standard:} 必要ないのにスマートフォンを持っている理由は緊急連絡用に親が\textbf{買ってくれた}から。\newline 
\textbf{English:} The reason I have a smartphone even though I don't need one is because my parents \textbf{bought it for me} as an emergency contact. \\
\bottomrule
\end{tabularx}
\vspace{-2mm}
\end{table*}

While Japanese demonstrated substantial improvements from both approaches, elevating the performance from the baseline of 24.21 to a maximum score of 28.26, German and Chinese exhibited contrasting behaviors regarding which method was effective. For German, data augmentation with pseudo-dialect speech successfully improved the dialect understanding performance compared to the augmentation with synthetic standard speech (+0.90 BLEU\rev{, $p < 0.001$}). Conversely, multi-task learning was not effective for German because direct translation is already effective. Forcing an intermediate text output likely caused error propagation rather than serving as a semantic bridge.
In contrast, for Chinese, pseudo-dialect speech augmentation showed no noticeable effect. However, multi-task learning with intermediate standard Mandarin prediction improved the dialect understanding performance from 11.67 to 16.37 (+4.70 BLEU). Chinese dialects share an ideogram-based written system with the standard language. Therefore, they share a strong grammatical and lexical foundation despite different pronunciations. Mapping the speech input to this standard text space thus likely functioned as an effective semantic bridge for the LLM backend. Conversely, this shared text does not alter its character forms to reflect phonetic variations. Unlike phonetic languages, reproducing dialect-specific readings using a standard TTS model is uniquely challenging. Consequently, the TTS model failed to synthesize distinct dialectal speech, potentially limiting the benefit of audio augmentation.
\rev{Furthermore, compared to the \texttt{Standard+English} setting, the \texttt{ASR+Standard+English} setting degraded performance across all three languages. This decline is likely attributed to the amplification of error propagation caused by passing through two intermediate predictions.}

We next examine the detailed performance across the 20 Japanese dialects\del{, where both proposed methods yielded synergistic gains}. As shown in Table~\ref{tab:data_augmentation_results}, pseudo-dialect augmentation improved the average BLEU score to 26.24, \rev{significantly} outperforming synthetic standard speech (25.38\rev{, $p < 0.001$}). \del{While it brought clear gains in 15 dialects, its effects were limited in the remaining 5 dialects (Iwaki, Izumo, Kanazawa, Nara, and Osaka) placed at the end of the table.} \rev{It brought gains over both the no-augmentation and synthetic standard speech baselines in 15 dialects (with 4 out of 20 dialects showing statistical significance, $p < 0.05$).}

Table~\ref{tab:multitask_results} demonstrates that incorporating the standard Japanese intermediate target (\texttt{Standard+English}) achieved further performance improvements\rev{, significantly outperforming the \texttt{ASR+English} target ($p < 0.05$)}. Compared to the \rev{direct} English translation, this configuration improved performance in 19 out of 20 dialects \rev{(with 9 dialects showing statistically significant improvements, $p < 0.05$)}, including \rev{dialects} where audio augmentation alone showed no gains. \del{As a result, it outperformed the baseline across all dialects.} In contrast, the ASR target (\texttt{ASR+English}) was less stable, showing larger performance fluctuations in specific dialects like \rev{Hiroshima,} Izumo\rev{,} and Tsugaru. This instability underscores that mapping to a standard text space provides a more robust semantic bridge than literal transcriptions.

\section{Analysis}\label{analysis}
This section provides a qualitative analysis to examine how our method improves dialect understanding and to identify remaining challenges. This evaluation is conducted using Japanese examples from Table~\ref{tab:qualitative_analysis} and Table~\ref{tab:multitask_comparison}, which showed clear performance improvements across both data augmentation and output format configurations. Through these cases, we demonstrate the specific advantages and limitations of pseudo-dialect speech augmentation, as well as the benefits of multi-task learning.

\subsection{Improvement by Pseudo-Dialect Speech}
As shown in the improvement examples, the proposed method is able to learn grammar and vocabulary unique to dialects. Specifically, the baseline model misrecognized the dialectal phrase ``たりひんかった(tarihinkatta)" as ``katta" that means ``bought" in Improved 1. Similarly, in Improved 2, it misrecognized the dialectal ending ``nappeyo'' (meaning ``will probably become'') as the phonetically similar standard expression ``natteyo'' (which expresses a desire or request).
In contrast, the proposed method correctly recognized the dialect expressions in both examples.
These results support that through learning with pseudo-dialect speech that has dialectal linguistic features even with standard language acoustic features, SLM can acquire knowledge regarding the grammar and vocabulary structure of dialects.

\subsection{Challenges of Pseudo-Dialect Speech}
On the other hand, the failure examples highlight two primary challenges. First is uncovered dialect vocabulary due to limitations in LLM translation quality. In failure example 1, both models failed to recognize the dialect term ``がめる(gameru)" that means ``pick up". Investigation of the training data revealed that all instances that should have been translated to ``がめる(gameru)" were missed by the LLM.
In failure example 2, both methods failed to recognize the standard word ``赤(aka)" that means ``red" due to the Tsugaru dialect accent. This indicates that the method has limitations in adapting to acoustic variations unique to dialects, since the pseudo-dialect speech is synthesized with standard language acoustic features.

\subsection{Improvement by Multi-task Learning}
Table \ref{tab:multitask_comparison} compares the outputs of the different models given a Tsugaru dialect input. The single-task model mistranslates the dialectal phrase ``わさけだ(wasakeda)", which means ``someone gave it to me." Furthermore, in the model jointly outputting transcriptions, the phrase is misrecognized as ``話さけだ," which is not an existing word sequence but could be pronounced similarly, and this error propagates directly to the English translation. In contrast, the model jointly outputting standard Japanese successfully captures the core semantic meaning of parental giving, translating the phrase as ``bought it for me.'' This configuration yields the most contextually accurate English output among the compared models. This indicates that standard Japanese serves as an effective intermediate representation that bridges the semantic gap between dialects and English.

\section{Effectiveness across different Model Architectures}\label{architectures}
To verify whether our proposed pseudo-dialect speech augmentation is effective beyond the SLM architecture, we also examine its impact on alternative frameworks, specifically direct English translation using Whisper, and a cascade speech translation system combining Whisper and Swallow. For these models, Whisper and Swallow are trained independently. Unlike the frozen configuration of the SLM, Whisper undergoes full-parameter fine-tuning with 1,543.5M total and trainable parameters, whereas Swallow is efficiently tuned using LoRA~\cite{hu2022lowrank} with 41.9M trainable parameters out of 8,072.2M total parameters. Consequently, the cascade system comprises 9,615.7M total parameters, with 1,585.4M trainable parameters. Due to these significant differences in trainable parameters, our evaluation focuses on the effectiveness of pseudo-dialect augmentation within each respective architecture rather than comparing absolute performance across them.

All models were trained on 8 NVIDIA A100 (80GB) GPUs for 10 epochs using the AdamW optimizer. \rev{For t}he individual training of Whisper and Swallow, the effective batch size was 256, and the peak learning rate was 1e-5. Training Whisper took about 1.5 days without augmentation and 3 days with it, while Swallow required about 1 day and 2 days, respectively.

The results of the English translation task using the SLM, Whisper, and the cascade system are presented in Table \ref{tab:architecture_comparison}. First, although the cascade system outperforms the SLM, this is primarily because both Whisper and Swallow underwent fine-tuning. Therefore, this result does not necessarily indicate the architectural superiority of one model over the other. Notably, not only for the SLM but also for both direct English translation using Whisper and the cascade translation system, applying the proposed pseudo-dialect speech augmentation improved translation performance compared to training with synthetic standard speech (+\rev{0.54} and +0.79 for Whisper and the cascade system, respectively\rev{, $p < 0.001$}). This demonstrates that, even after isolating the effect of simply increasing the data volume, the proposed pseudo-dialect speech augmentation is generally effective across various architectures.

\begin{table}[t]
  \caption{Effectiveness of pseudo-dialect augmentation across architectures in English translation (BLEU). ``No training'' means no fine-tuning. ``+ Standard'' and ``+ Pseudo-dialect'' add synthetic standard and pseudo-dialect speech. SLM scores match averages in Table~\ref{tab:data_augmentation_results}.}
  \vspace{-2mm}
  \label{tab:architecture_comparison}
  \small
  \centering
  \begin{tabular}{lccc}
    \toprule
    \textbf{Dataset} & \textbf{SLM} & \textbf{Whisper} & \textbf{Cascade} \\
    \midrule
    \# Total params & 8,683.7M & 1,543.5M & 9,615.7M \\
    \# Trainable params & 18.4M & 1,543.5M & 1,585.4M \\
    \midrule
    No training & - & 14.36 & 19.87 \\
    No Augmentation & 24.21 & 17.78 & 29.78 \\
    + Standard & 25.38 & 18.44 & 29.65 \\
    + Pseudo-dialect & \textbf{26.24} & \textbf{18.98} & \textbf{30.44} \\
    \bottomrule
  \end{tabular}
  \vspace{-4mm}
\end{table}

\section{Conclusion}
In this study, we proposed dialect adaptation for SLM using pseudo-dialect speech synthesized without real dialect speech, combined with multi-task learning leveraging the corresponding standard language text. Experimental results demonstrated that our method \del{consistently }improves dialect understanding performance across multiple languages and alternative model architectures. This serves as a practical and scalable strategy to expand the dialect coverage of SLM, eliminating the need for extensive dialect speech collection.

\del{While our quantitative results validated the effectiveness of pseudo-dialect speech, the qualitative analysis in Section V revealed the essential need to generate data that possesses actual dialectal phonetic features rather than standard acoustic features. However, just as with text data, collecting authentic speech for various dialects remains highly challenging. Therefore, our future work will focus on developing and integrating advanced voice conversion or speech synthesis technologies that can replicate these realistic dialectal features under low-resource constraints.}

\rev{The qualitative analysis in Section V illustrated how the proposed method improves dialect understanding, while also highlighting the importance of LLM dialect translation quality and cases where the TTS system needs to capture dialectal accents. Therefore, establishing evaluation protocols for low-resource dialect text translation, as well as developing advanced speech synthesis technologies that replicate realistic dialectal features under low-resource constraints, remain key directions for future work.}



\newpage
\bibliographystyle{IEEEtran}
\bibliography{mybib}

\end{document}